\documentclass{article} 
\usepackage{configuration/iclr2026_conference,times}

\usepackage{hyperref}
\usepackage{url}

\title{Beyond Model-centric: Collaborative DATA Optimization for Reusing and Sharing}
\author{Xinyi Shang$^{1,\ast}$ \quad Peng Sun$^{2,3} \thanks{Equal contribution with the first author}$\quad 
Fengyuan Liu $^{4}$ \quad Tao Lin$^{3,}\thanks{Corresponding author.}$\\
    $^{1}$University College London \quad $^{2}$Zhejiang University 
    \quad $^{3}$Westlake University\\
    $^{4}$University of Science and Technology of China\\
    {\tt\small xinyi.shang.23@ucl.ac.uk, sunpeng@westlake.edu.cn}\\
    {\tt\small lfy1231@mail.ustc.edu.cn, lintao@westlake.edu.cn}
}

\iclrfinalcopy 

\usepackage{graphicx} 
\usepackage{wrapfig}
\usepackage{array}

\usepackage{amsmath,amssymb,amsthm}

\newcommand{\cry}{\raisebox{-0.5ex}{\includegraphics[height=1.1em]{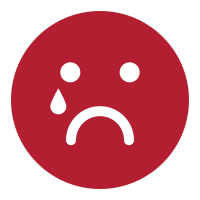}}\xspace}

\newcommand{\ssmile}{\raisebox{-0.5ex}{\includegraphics[height=1.05em]{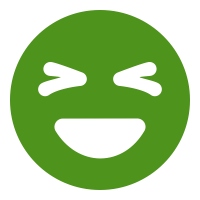}}\xspace}

\newcommand{\algopt}{\textsc{CoOpt}\xspace}
\newcommand*\circled[1]{\tikz[baseline=(char.base)]{\node[shape=circle,draw,inner sep=.3pt] (char) {#1};}}

\definecolor{orange}{RGB}{178,92,35}
\definecolor{green1}{RGB}{95,145,51}

\definecolor{red1}{RGB}{197,64,57}
\definecolor{blue1}{RGB}{59,130,220}
\definecolor{green2}{RGB}{82,181,150}
\definecolor{purple1}{RGB}{105,93,223}
\definecolor{orange1}{RGB}{164, 47, 54}
\definecolor{green3}{RGB}{94,145,51}

\definecolor{c_step1}{RGB}{60, 105, 199}
\definecolor{c_step2}{RGB}{159,37,28}
\definecolor{c_step3}{RGB}{107,40,157}

\providecommand{\mpsi}{\boldsymbol{\psi}}

\providecommand{\R}{\mathbb{R}} %

\DeclareMathOperator*{\argmin}{arg\,min\,}

\providecommand{\tt}{\mathbf{t}}

\providecommand{\xx}{\mathbf{x}}
\providecommand{\yy}{\mathbf{y}}

\providecommand{\mD}{\mathbf{D}}

\providecommand{\mS}{\mathbf{S}}
\providecommand{\mT}{\mathbf{T}}

\providecommand{\mW}{\mathbf{W}}

\providecommand{\mphi}{\boldsymbol{\phi}}

\providecommand{\mtheta}{\boldsymbol{\theta}}

\providecommand{\cT}{\mathcal{T}}

\providecommand{\cV}{\mathcal{V}}

\newenvironment{talign*}
{\csname align*\endcsname}
{\endalign}

\usepackage[utf8]{inputenc}         %
\usepackage[T1]{fontenc}            %
\usepackage{url}                    %
\usepackage{booktabs}               %
\usepackage{amsfonts}               %
\usepackage{nicefrac}               %
\usepackage{microtype}              %
\usepackage{xcolor}                 %
\usepackage{algorithm}
\usepackage{algorithmic}
\usepackage{graphicx}
\usepackage{subcaption}
\usepackage[flushleft]{threeparttable}
\usepackage{float}
\usepackage{multirow}
\usepackage{xspace}
\usepackage{natbib}
\usepackage{enumitem}
\usepackage[font=small]{caption}
\usepackage{autobreak}
\usepackage{sidecap}
\usepackage{wrapfig}
\usepackage{bbding}
\usepackage[toc, page, header]{appendix}
\usepackage{tikz}
\usepackage{xcolor}
\usepackage{pifont}
\usepackage{mdframed}
\usepackage{colortbl}

\hypersetup{
  colorlinks=true,
  linkcolor=blue,
  citecolor=blue,
  urlcolor=blue
}

\definecolor{coral}{RGB}{255,127,80}
\definecolor{darkgreen}{RGB}{0,100,0}
\definecolor{darkyellow}{RGB}{204,153,0}
\definecolor{salmon}{RGB}{250,128,114}
\definecolor{darkred}{RGB}{150,0,0}
\newcommand{\darkredtext}[1]{{\color{darkred}#1}}

\newcommand{\secref}[1]{\hyperref[#1]{\darkredtext{Sec.~\ref*{#1}}}}
\newcommand{\thmref}[1]{\hyperref[#1]{\darkredtext{Thm.~\ref*{#1}}}}
\newcommand{\defref}[1]{\hyperref[#1]{\darkredtext{Def.~\ref*{#1}}}}
\newcommand{\propref}[1]{\hyperref[#1]{\darkredtext{Prop.~\ref*{#1}}}}
\newcommand{\assumpref}[1]{\hyperref[#1]{\darkredtext{Assump.~\ref*{#1}}}}
\newcommand{\remarkref}[1]{\hyperref[#1]{\darkredtext{Rem.~\ref*{#1}}}}
\newcommand{\hypref}[1]{\hyperref[#1]{\darkredtext{Hyp.~\ref*{#1}}}}
\newcommand{\conjref}[1]{\hyperref[#1]{\darkredtext{Conj.~\ref*{#1}}}}
\newcommand{\lemref}[1]{\hyperref[#1]{\darkredtext{Lem.~\ref*{#1}}}}
\newcommand{\corref}[1]{\hyperref[#1]{\darkredtext{Cor.~\ref*{#1}}}}
\newcommand{\noteref}[1]{\hyperref[#1]{\darkredtext{Nota.~\ref*{#1}}}}
\newcommand{\claimref}[1]{\hyperref[#1]{\darkredtext{Clm.~\ref*{#1}}}}
\newcommand{\obsref}[1]{\hyperref[#1]{\darkredtext{Obs.~\ref*{#1}}}}
\newcommand{\algref}[1]{\hyperref[#1]{\darkredtext{Alg.~\ref*{#1}}}}
\newcommand{\figref}[1]{\hyperref[#1]{\darkredtext{Fig.~\ref*{#1}}}}
\newcommand{\tabref}[1]{\hyperref[#1]{\darkredtext{Tab.~\ref*{#1}}}}
\newcommand{\appref}[1]{\hyperref[#1]{\darkredtext{App.~\ref*{#1}}}}

\newtheoremstyle{custom}
{1pt} 
{1pt} 
{\itshape} 
{} 
{\bfseries} 
{} 
{ } 
{\thmname{#1} \thmnumber{#2} \thmnote{(#3)} . } 

\theoremstyle{custom}

\newtheorem{innerdefinition}{Definition}
\newtheorem{innerproposition}{Proposition}
\newtheorem{innerassumption}{Assumption}
\newtheorem{innerremark}{Remark}
\newtheorem{innertheorem}{Theorem}
\newtheorem{innerhypothesis}{Hypothesis}
\newtheorem{innerconjecture}{Conjecture}
\newtheorem{innerlemma}{Lemma}
\newtheorem{innercorollary}{Corollary}
\newtheorem{innernotation}{Notation}
\newtheorem{innerclaim}{Claim}
\newtheorem{innerproblem}{Problem}

\newtheorem{innerobservation}{Observation}

\newmdenv[
  backgroundcolor=gray!10,
  linecolor=gray!100,
  linewidth=0.8pt,
  skipabove=2pt,
  skipbelow=2pt,
  innertopmargin=10pt,
  innerbottommargin=5pt,
  innerleftmargin=5pt,
  innerrightmargin=5pt,
]{definitionframe}

\newmdenv[
  backgroundcolor=blue!10,
  linecolor=blue!100,
  linewidth=0.8pt,
  skipabove=2pt,
  skipbelow=2pt,
  innertopmargin=10pt,
  innerbottommargin=5pt,
  innerleftmargin=5pt,
  innerrightmargin=5pt,
]{propositionframe}

\newmdenv[
  backgroundcolor=green!10,
  linecolor=green!100,
  linewidth=0.8pt,
  skipabove=2pt,
  skipbelow=2pt,
  innertopmargin=10pt,
  innerbottommargin=5pt,
  innerleftmargin=5pt,
  innerrightmargin=5pt,
]{assumptionframe}

\newmdenv[
  backgroundcolor=yellow!10,
  linecolor=yellow!100,
  linewidth=0.8pt,
  skipabove=2pt,
  skipbelow=2pt,
  innertopmargin=10pt,
  innerbottommargin=5pt,
  innerleftmargin=5pt,
  innerrightmargin=5pt,
]{remarkframe}

\newmdenv[
  backgroundcolor=red!10,
  linecolor=red!100,
  linewidth=0.8pt,
  skipabove=2pt,
  skipbelow=2pt,
  innertopmargin=10pt,
  innerbottommargin=5pt,
  innerleftmargin=5pt,
  innerrightmargin=5pt,
]{theoremframe}

\newmdenv[
  backgroundcolor=purple!10,
  linecolor=purple!100,
  linewidth=0.8pt,
  skipabove=2pt,
  skipbelow=2pt,
  innertopmargin=10pt,
  innerbottommargin=5pt,
  innerleftmargin=5pt,
  innerrightmargin=5pt,
]{hypothesisframe}

\newmdenv[
  backgroundcolor=orange!10,
  linecolor=orange!100,
  linewidth=0.8pt,
  skipabove=2pt,
  skipbelow=2pt,
  innertopmargin=10pt,
  innerbottommargin=5pt,
  innerleftmargin=5pt,
  innerrightmargin=5pt,
]{conjectureframe}

\newmdenv[
  backgroundcolor=cyan!10,
  linecolor=cyan!100,
  linewidth=0.8pt,
  skipabove=2pt,
  skipbelow=2pt,
  innertopmargin=10pt,
  innerbottommargin=5pt,
  innerleftmargin=5pt,
  innerrightmargin=5pt,
]{lemmaframe}

\newmdenv[
  backgroundcolor=magenta!10,
  linecolor=magenta!100,
  linewidth=0.8pt,
  skipabove=2pt,
  skipbelow=2pt,
  innertopmargin=10pt,
  innerbottommargin=5pt,
  innerleftmargin=5pt,
  innerrightmargin=5pt,
]{corollaryframe}

\newmdenv[
  backgroundcolor=pink!10,
  linecolor=pink!100,
  linewidth=0.8pt,
  skipabove=2pt,
  skipbelow=2pt,
  innertopmargin=10pt,
  innerbottommargin=5pt,
  innerleftmargin=5pt,
  innerrightmargin=5pt,
]{notationframe}

\newmdenv[
  backgroundcolor=violet!10,
  linecolor=violet!100,
  linewidth=0.8pt,
  skipabove=2pt,
  skipbelow=2pt,
  innertopmargin=10pt,
  innerbottommargin=5pt,
  innerleftmargin=5pt,
  innerrightmargin=5pt,
]{claimframe}

\newmdenv[
  backgroundcolor=salmon!10,
  linecolor=salmon!100,
  linewidth=0.8pt,
  skipabove=2pt,
  skipbelow=2pt,
  innertopmargin=10pt,
  innerbottommargin=5pt,
  innerleftmargin=5pt,
  innerrightmargin=5pt,
]{problemframe}

\newmdenv[
  backgroundcolor=lavender!10,
  linecolor=lavender!100,
  linewidth=0.8pt,
  skipabove=2pt,
  skipbelow=2pt,
  innertopmargin=10pt,
  innerbottommargin=5pt,
  innerleftmargin=5pt,
  innerrightmargin=5pt,
]{observationframe}

\newenvironment{definition}
{\begin{definitionframe}\begin{innerdefinition}}
      {\end{innerdefinition}\end{definitionframe}}

\usepackage[textwidth=2.6cm,textsize=tiny]{todonotes}

\begin{document}

\maketitle

\begin{abstract}

    This paper pioneers a \textit{novel data-centric paradigm} to maximize the utility of unlabeled data, tackling a critical question: \emph{How can we enhance the sustainability and efficiency of deep learning training by optimizing the data itself?}
    We begin by identifying two key limitations in existing model-centric approaches, all rooted in a shared bottleneck: knowledge extracted from data is locked to model parameters, hindering its reusability and scalability. 
    To this end, we propose \algopt, a highly efficient, parallelized framework for collaborative unlabeled data optimization.
    By distributing unlabeled data and leveraging publicly available task-agnostic prior models, \algopt optimizes raw unlabeled data into knowledge-enriched training sets that are effective, efficient, reusable, and easily shareable.   
    Extensive experiments across diverse datasets and architectures validate these advantages, achieving a 7.9\% improvement on ImageNet-1K over BYOL. 
    Notably, \algopt remains effective even when all prior models are significantly weak, substantially accelerating the early stages of training. These results establish data-centric optimization as a promising path toward sustainable and efficient deep learning \footnote{Code will be publicly available.}.
\end{abstract}

\section{Introduction}\label{sec:intro}

\begin{wrapfigure}{r}{0.46\textwidth}
    \vspace{-2.5em}
    \centering
    \includegraphics[width=.46\textwidth]{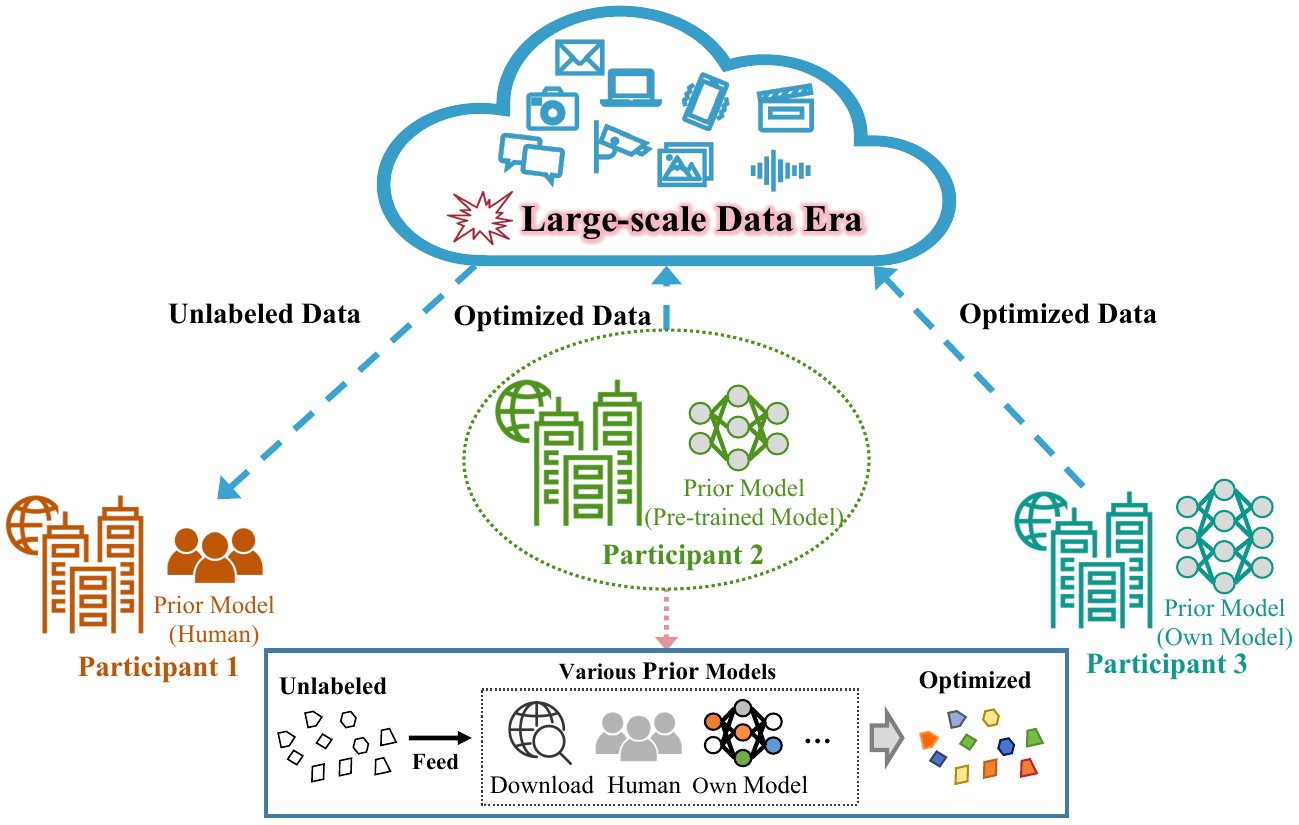}
    \vspace{-1em}
    \caption{
        \textbf{A \underline{Co}llaborative Data \underline{Opt}imization Framework \algopt.}
        For large-scale unlabeled data, self-supervised learning results in \textcolor{orange1}{\cry low training efficiency}.
        Therefore, we propose \algopt, an \textcolor{green3}{\ssmile efficient and parallel framework} enabling participants to use diverse task-agnostic models, such as pre-trained ResNets, termed \textit{prior models}, for collaborative data optimization.
    }
    \label{fig:overview}
    \vspace{-1.2em}
\end{wrapfigure}



Deep Learning has achieved remarkable success, primarily due to the large-scale datasets \citep{Song2020Learning,Yang2023Interactive}.
Despite the abundance of data in the era of big data, a significant portion of them remains unlabeled \citep{lei2023comprehensive}.
The dominant paradigm in the field for exploiting unlabeled data is self-supervised learning (SSL), which is fundamentally \emph{model-centric}: it carefully crafted pretext tasks and loss functions to encode data information into model parameters \citep{chen2020simple, grill2020bootstrap,gui2024survey}.

However, the model-centric nature presents two critical challenges.
\textbf{\emph{First}}, their training protocols are tightly coupled to specific network architectures, severely hindering the transferability and reusability of trained model to other architectures~\citep{wagner2022importance,huang2023self}.
\textbf{\emph{Second}}, despite acceleration advances, training over extensive unlabeled datasets still computationally prohibitive~\citep{sun2024efficiency}.
At the core of these challenges is a shared bottleneck: knowledge extracted from data is locked in model parameters, restricting its adaptability and preventing efficient reuse across diverse tasks or architectures.

To break free from model-centric paradigm, we propose a data-centric paradigm that directly optimizes the unlabeled data by optimizing targets for samples (detailed in \secref{sec:data_opt}), thereby effectively encoding knowledge into the data itself rather than into model parameters.
The resulting ``optimal data'' is agnostic to downstream architectures, accelerates subsequent training by providing richer supervision, and can be reused across multiple tasks without repeated large-scale pretraining.

\begin{figure*}[!h]
    \centering
    \vspace{-5pt}
    \includegraphics[width=\textwidth]{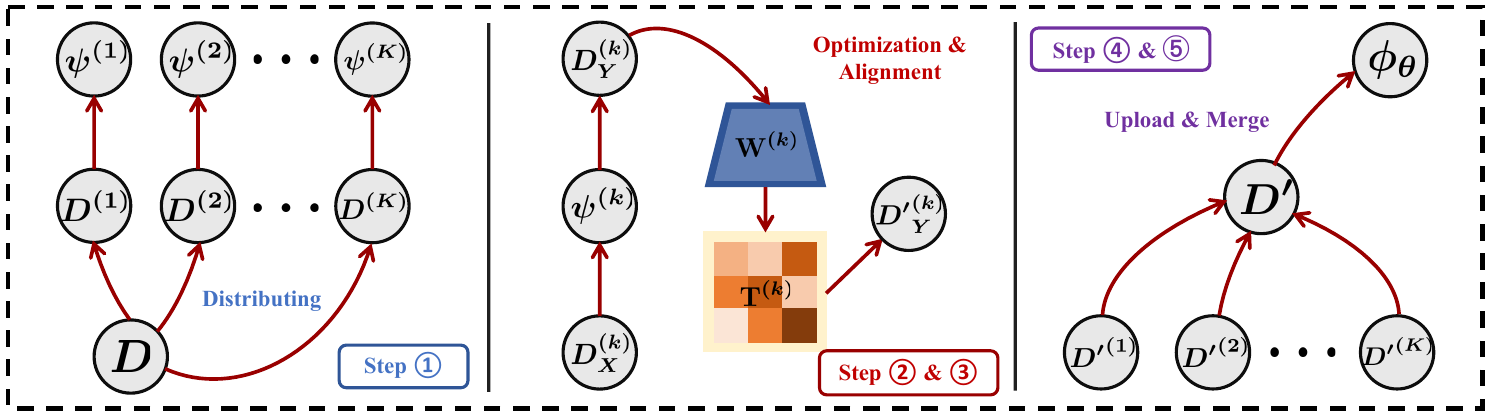}\\
    \caption{\textbf{Lifecycle of the proposed collaborative data optimization framework \algopt}.
        The framework encompasses an open data platform and multiple participants, involving five key data operations.}
    \label{fig:process}
    \vspace{-10pt}
\end{figure*}

In the meanwhile, scaling this approach to massive unlabeled datasets introduces a significant challenge: a single node faces prohibitive compute and storage demands.
To address this, we propose \algopt, a highly efficient collaborative framework inspired by crowd-sourcing to achieve parallel data optimization.
An overview of \algopt is depicted in \figref{fig:overview}, with detailed processes shown in \figref{fig:process}.
In \algopt, the unlabeled dataset is partitioned into disjoint subsets, each processed independently and in parallel by participants equipped with task-agnostic models.
These models, referred to as \textit{prior models}, can diverge from publicly available pre-trained models and the participants' local models.
Once the targets are optimized, they are aggregated to reconstruct a fully optimized dataset, achieving computational scalability through decentralized workload distribution.

\algopt offers \textit{\textbf{three key advantages}}:
First, relying solely on task-agnostic prior models, the optimized data can be directly used to any downstream architecture, thereby ensuring strong generalizability and reusability (see \secref{sec:exp_sota}).
Second, by distributing non-overlapping data subsets across participants, each node handles only a fraction of the total computational cost, thereby enabling scalable and resource-efficient optimization (see \secref{sec:exp_sota}).
Third, \algopt is lightweight, incurring negligible overhead while substantially enhancing training efficiency and performance (see \secref{sec:exp_ablation}).

\textbf{In summary, our contributions are threefold:}
\begin{enumerate}[label=(\alph*), nosep, leftmargin=16pt, itemsep=3pt]
    \item We propose \algopt, the \textit{first} data-centric framework for collaboratively optimizing unlabeled data. 
          By leveraging task-agnostic prior models, \algopt transforms raw unlabeled samples into optimal data, enabling high performance, efficiency, strong generalization, and reusability.
    \item Within \algopt, we identify a critical issue, \textit{Target Distribution Inconsistency} (\secref{sec:definition_issue}), and introduce a lightweight target alignment strategy to address it (\secref{sec:uniform_value}). 
    \item We conduct experiments across datasets and models to comprehensively validate the advantages of \algopt (\secref{sec:exp_sota}).
Further, we provide a detailed analysis of the key factors influencing its effectiveness (\secref{sec:exp_factor}).
          Remarkably, we demonstrate that \algopt remains effective even when all prior models are weak, substantially accelerating the early stages of training.
\end{enumerate}



\vspace{-10pt}
\section{Related Work}
\label{sec:related_work}

\vspace{-5pt}
\paragraph{Self-Supervised Learning.}
It aims to exploit the intrinsic relationships within unlabeled data.
For example, InstDisc \citep{wu2018unsupervised} uses instance discrimination as a pretext task.
MoCo \citep{he2020momentum} significantly increases the number of negative samples but uses a simplistic strategy for selecting positive samples.
SimCLR \citep{chen2020simple} highlights the importance of hard positive sample strategies.
Notably, BYOL \citep{grill2020bootstrap} discards negative sampling and surpasses the performance of SimCLR \citep{chen2020simple}.


\vspace{-8pt}
\paragraph{Model-Centric Perspective: Knowledge Distillation.}
Knowledge distillation \citep{hinton2015distilling} leverages teacher-generated soft labels to improve student training efficiency and performance \citep{Dong2023DisWOT:}.
A line of knowledge distillation methods utilizes multiple teachers (MKD) \citep{zhang2022confidence,pham2023collaborative} to enhance student learning.
They assume that ensemble outputs from multiple teachers enables students to learn more generalized representations.
Notably, all teacher models process the same input data. 

\begin{wrapfigure}{r}{0.6\textwidth}
    \vspace{-2em}
    \centering
    \includegraphics[width=.6\textwidth]{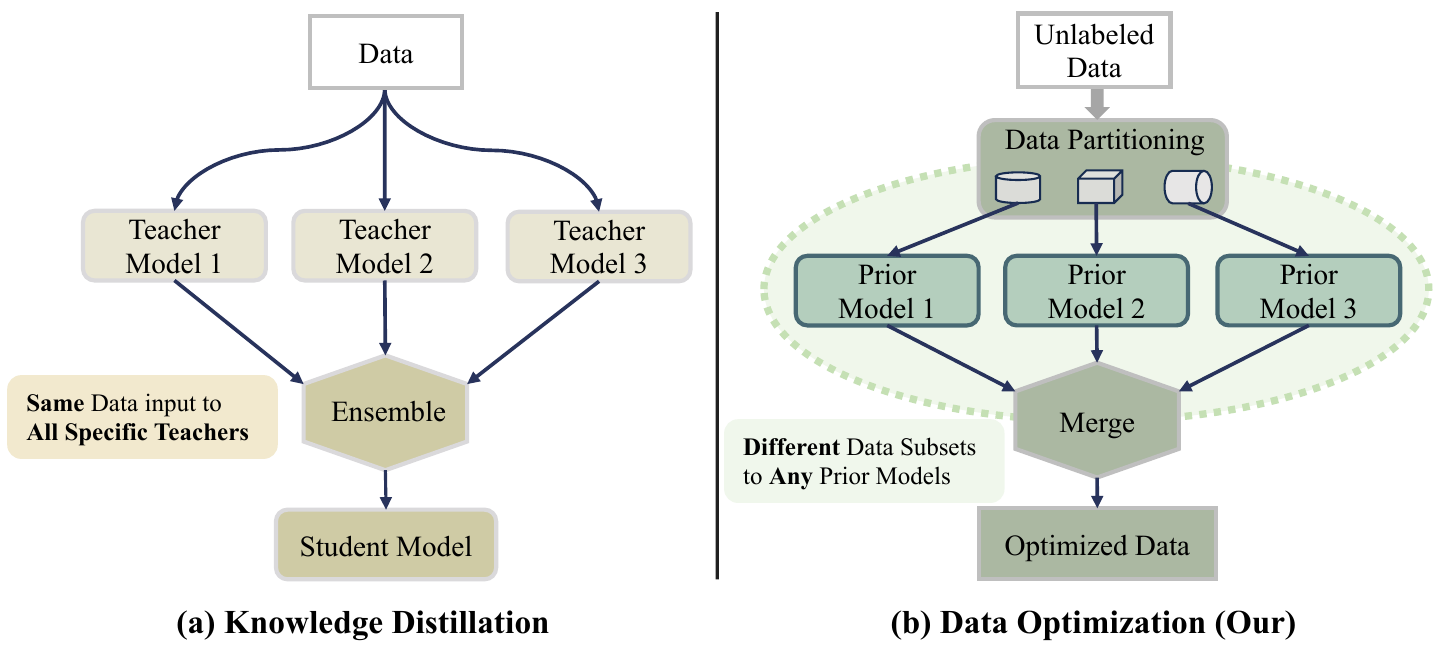}
    \vspace{-1.2em}
    \caption{
       Comparison Between KD and Ours.
    }
    \label{fig:kd_ours}
    \vspace{-1em}
\end{wrapfigure}

\textit{Our setting departs fundamentally from knowledge distillation in terms of objective, input data, and teacher models (see \figref{fig:kd_ours}).}
First, in KD, the distilled knowledge is embedded in student parameters, limiting its reuse across different architectures.
In contrast, our objective is to construct a high-quality, optimized dataset that is model-agnostic and reusable, enabling training or evaluation of diverse architectures.
Second, rather than feeding all teachers the same inputs, we partition the unlabeled data into disjoint subsets, each optimized by a different prior model.
Third, existing KD methods often rely on intricate loss functions \citep{jiang2024mtkd} or require teacher fine-tuning \citep{wu2021one}, but our framework uses arbitrary pre-trained models without domain-specific adaptation.
The optimized data can then be directly reused to train arbitrary downstream models without further modification.

\vspace{-8pt}
\paragraph{Data-Centric Perspective: Dataset Distillation.}
Dataset distillation \citep{wang2018dataset} improves the training efficiency by learning a compact distilled dataset that can achieve comparable performance to the original dataset with less training cost.
The majority of methods focus on optimizing images, which can be categorized into three primary approaches \citep{lei2023comprehensive}: meta-learning frameworks \citep{wang2018dataset, zhou2022dataset}, matching-based methods \citep{zhao2020dataset, zhao2023dataset, guo2023towards} and decoupling frameworks \citep{yin2023squeeze, sun2024efficiency}.
Notably, current methods predominantly focus on distilling labeled datasets.

\vspace{-5pt}
\section{Collaborative Data Optimization Framework \algopt}
\vspace{-5pt}
We begin by formally defining \textit{data optimization} in \secref{sec:data_opt}.
Subsequently, we provide a detailed description of the proposed \algopt in \secref{sec:framework}.
Furthermore, we identify an inherent challenge within this framework in \secref{sec:definition_issue} and present method in \secref{sec:uniform_value}.

\subsection{Definition of Data Optimization}\label{sec:data_opt}
\vspace{-5pt}
We first revisit current training acceleration techniques, including knowledge distillation \citep{hinton2015distilling} and dataset distillation \citep{wang2018dataset}.
Specifically, we decouple data into samples $D_X$ and targets $D_Y$. 
Essentially, compared to self-supervised learning methods, these approaches are more efficient.
They achieve this by either optimizing the target $D_Y$ of pre-trained models (as in knowledge distillation) or by jointly optimizing both the input data $D_X$ and target $D_Y$ (as in dataset distillation).
Notably, for dataset distillation, optimizing input data $D_X$ is computationally more expensive than optimizing targets $D_Y$ \citep{bohdal2020flexible}.
Furthermore, recent studies \cite{shang2025gift,qin2024label} have indicated that solely optimizing $D_Y$ not only reduces computational overhead but also achieves significant performance gains.
These findings have demonstrated that \textit{optimizing $D_Y$ is both necessary and efficient.}
We refer to this process as \textbf{\emph{data optimization}}.

Formally, consider a large-scale unlabeled dataset $D = D_X = \{\xx_i\}_{i=1}^N$, where $\xx_i \in \R^m$ and $N=|D|$, \emph{data optimization} aims to assign targets $D_Y=\{\yy_i\}_{i=1}^{N}$ to construct an optimally labeled dataset
$D^\prime = \{ (\xx_i, \yy_i)\}_{i=1}^{N}$ such that models trained on $D^\prime $ can achieve \textbf{\textit{higher}} performance of those trained on $D$ with \textbf{\textit{significantly less training costs}}.
This objective is expressed as
\begin{equation}\label{eq:define_dd}
    \textstyle
    \mathbb{E}_{(\xx, y) \sim P_\cT} [ \ell ( \mphi_{\mtheta_D}( \xx ), y ) ] > \mathbb{E}_{(\xx, y) \sim P_\cT} [ \ell ( \mphi_{\mtheta_{D^\prime }}( \xx ), y ) ] \,,
\end{equation}
where $P_\cT$ denotes the test distribution, $\mathbf{x}$ is a test sample, $y$ is its label, $\ell$ is the loss function (e.g., cross-entropy loss), and $\mtheta_D$ and $\mtheta_{D^\prime}$ are parameters of network $\mphi$ trained on $D$ and $D^\prime$, respectively.

Notably, directly extending these methods to unlabeled data is infeasible. 
Existing dataset distillation methods focus on labeled data, whereas knowledge distillation methods, as reviewed in \secref{sec:related_work}, significantly differ from ours. 
A detailed comparison is presented in \figref{fig:kd_ours}.
To address these limitations, we propose a collaborative framework that leverages distributed computation and various task-agnostic models for unlabeled data.
Specifically, inspired by \citep{sun2024efficiency}, which shows that using task-agnostic models for target assignment can expedite training, we further enhance efficiency by splitting the data and then parallelly optimizing each split.
After optimization, the optimal subset are aggregated to reconstruct a fully optimized dataset, achieving computational efficiency.
Formally, we define data optimization with a prior model $\mpsi$ in each participant in \defref{pro:1}.
\begin{definition}[Data optimization with prior model $\mpsi$]\label{pro:1}
    Given samples $D_X = \{\xx_i\}_{i=1}^{N}$ and a prior model $\mpsi: \R^m \to \R^l$, data optimization assigns optimal targets $D_Y = \{\yy_i\}_{i=1}^{N}$ for the samples to create $D^\prime = \{\xx_i, \yy_i\}_{i=1}^{N}$.
    We assigns a target $\yy_i$ for $\xx_i$ as:
    \begin{equation}
        \textstyle
        D^\prime  = \{ (\xx_i,\yy_i) \mid \yy_i = \mW \mpsi(\xx_i), \forall \xx_i \in D_X\} \,,
    \end{equation}
    where $\yy_i$ is the optimized target, and $\mpsi(\xx_i)$ represents the target of $\xx_i$, which means the feature representation.
    $\mW: \R^l \to \R^n$ denotes a matrix designed to transform the feature vector $\mpsi(\xx_i)$ from dimension $l$ to $n$ without loss of information \citep{matouvsek2008variants}.
    This transformation aligns the output dimension\footnote{Here, $n$ is the target dimensionality of $\mphi_{\mtheta_{D^\prime}}$.
        In practice, each participant produces targets of varying dimensions due to using different prior models.
        Therefore, to train the model $\mphi_{\mtheta_{D^\prime}}$ on the optimized data $D^\prime$, we employ a random matrix $\mW$ to transform all target vectors to a common dimensionality.} with that required by the model trained on optimized data $\mphi_{\mtheta_{D^\prime}}: \R^m \to \R^n$.
\end{definition}


\vspace{-10pt}
\subsection{Overview of the Proposed Framework \algopt}\label{sec:framework}
\algopt is a collaborative and parallelized framework that includes an open data platform and $K$ participants, each equipped with a distinct prior model.
\algopt operates through the five steps:

\textcolor{c_step1}{\textbf{Step \circled{1}: Data Distributing.}}
The open data platform initiates the process by randomly partitioning the entire set of unlabeled data $D$ into $K$ non-overlapping subsets.
Each participant then downloads one of these subsets from the platform, denoted as $D^{(k)}$, where $k$ denotes the $k$-th participant. 

\vspace{-2pt}
\textcolor{c_step2}{\textbf{Step \circled{2}: Data Optimization.}}
Each participant optimizes their respective unlabeled dataset $D^{(k)}$ using the prior model $\mpsi^k$.
This data optimization process, defined in \propref{pro:1}, yields optimized targets $D^{(k)}_Y$ and optimized data ${D^\prime}^{(k)}$.

\vspace{-2pt}
\textcolor{c_step2}{\textbf{Step \circled{3}: Data Alignment.}}
The heterogeneity of prior models among participants induces significant variations in their distribution of the targets.
This issue, referred to as \emph{target distribution inconsistency} (defined in \secref{sec:definition_issue}), necessitates an alignment strategy (detailed in \secref{sec:uniform_value}) to align the target distribution across participants.
Crucially, each participant needs to align their targets distribution to the most optimal prior model using a learnable transformation matrix $\mT^{(k)}$, yielding the final optimized dataset ${D^\prime}^{(k)}$.
Further details are provided in \secref{sec:uniform_value}.

\vspace{-2pt}
\textcolor{c_step3}{\textbf{Step \circled{4}: Data Uploading.}}
After optimization and alignment, participants upload their optimized datasets $\{{D^\prime}^{(k)}\}_{k=1}^K$ back to the open data platform.

\vspace{-2pt}
\textcolor{c_step3}{\textbf{Step \circled{5}: Data Merging.}}
The platform aggregates all the optimized datasets received from the participants to form a consolidated dataset.

The proposed \algopt enables participants to independently and parallelly optimize their subsets while ensuring consistency through the proposed alignment strategies.
Consequently, this approach markedly reduces individual data optimization costs and enhances efficiency.

\vspace{-5pt}
\subsection{An Inherent Challenge: Target Distribution Inconsistency}\label{sec:definition_issue}
In our collaborative framework, each participant may employ a distinct prior model, leading to inconsistencies in the target distributions, as illustrated in \figref{fig:vis_inconsistency_before}.
For example, participant 1 uses ResNet-18 for optimization, resulting in a target dimension of $512$, while participant 2 utilizes ResNet-50, yielding a target dimension of $2{,}048$.
Such inconsistencies can negatively impact the generalization capabilities of models trained on the optimized data, as they prevent the models from learning representations that are uniformly representative of the overall data distribution.



\vspace{-5pt}
\subsection{An Effective Strategy: Target Alignment}\label{sec:uniform_value}
To address this issue, a potential solution is to align the target distributions of all participants' prior models with that of the prior model producing the most optimal target distribution, referred to as the \textit{best prior model}.
Such alignment can be achieved by utilizing an optimizable transformation matrix to map each participant's target distribution to that of the best prior model \citep{sun2024efficiency}.
This alignment strategy ensures consistency across all optimized target distributions.

In summary, \textit{it is crucial to first effectively assess each participant's prior model quality and subsequently train the transformation matrix for alignment.}

\paragraph{A Metric to Quantify Prior Model Quality.}
Inspired by \citet{wang2020understanding}, which proposes an optimizable metric a.k.a.\ \emph{uniform value loss} to achieve feature uniformity on the hypersphere during training, we employ this metric to evaluate the quality of prior models.
Notably, \citep{wang2020understanding} also provide \textit{theoretical validation} of the connection between uniformity and feature quality.
Specifically, each participant downloads a small shared dataset $S_X$ from the platform and computes the uniformity value of their prior model on $S_X$.
They then upload this value to the platform, enabling it to determine which participant possesses the best prior model.
The uniform value is computed as:
\begin{equation}
    \textstyle
    \cV_{\text{uniform}}(\mpsi; S) \triangleq \log \mathbb{E}_{\xx_i, \xx_j \sim S} \left[ e^{\tau \|{\mpsi}(\xx_i) - {\mpsi}(\xx_j)\|_2^2} \right],
\end{equation}
where $\mpsi$ is the prior model, $\tau$ is a hyper-parameter set as $2$, consistent with~\citep{wang2020understanding}.

A lower uniform value indicates a higher-quality prior model, which optimizes targets of superior quality.
Extensive experiments in \figref{fig:ablation_uniform} demonstrate a strong correlation between this metric and the performance of prior, thereby effectively assessing the quality of targets.

\vspace{-5pt}
\paragraph{Alignment.}
Upon identifying the best prior model, denoted as $\mpsi^\star$, all participants, excluding the best prior model itself, proceed to train an optimizable transformation matrix.
Specifically, the participant owning $\mpsi^\star$ computes its optimized targets, denoted as ${\mS_Y}^\star$, on the shared dataset $S_X$.
${\mS_Y}^\star$ are then uploaded to the platform, which ensures they are publicly accessible to all participants.
Following this, each participant $k$ optimizes a lightweight transformation matrix, denoted as $\mT^{(k)}$, on the shared dataset $S_X$.
The optimization problem is defined as follows:
\begin{equation}
    \textstyle
    \mT^{(k)} = \argmin_{\mT \in \R^{n \times n} }  \{ \| \mT \cdot {\mpsi}^{(k)}(\mS_X) - {\mS_Y}^{\star} \|_2^2 \} \,,
\end{equation}
where $\mS_X$ represents the matrix form of $S_X$, suitable for input into the network ${\mpsi}^{(k)}$, and ${\mS_Y}^{\star}$ also represents the matrix form of ${S_Y}^{\star}$.
After obtaining the transformation matrix $\mT^{(k)}$, the participant can convert the optimized targets for its own data using this matrix:
${D_Y}^{(k)} = \mT^{(k)} \cdot {\mpsi}^{(k)}({\mD_X}^{(k)})$,
where ${\mD_X}^{(k)}$ denotes the participant's subset, and ${D_Y}^{(k)}$ are the adjusted targets aligned with the best prior model's target distribution.
As illustrated in \figref{fig:vis_inconsistency_after_bc}, the proposed alignment strategy effectively mitigates target distribution inconsistency.



\vspace{-5pt}
\paragraph{Remark on Privacy.}
Notably, this work focuses on optimizing large-scale open-source unlabeled data obtained from publicly available sources, such as the internet.
The information transmitted between the platform and participants is targets generated by prior models, which ensures that no direct privacy-sensitive information is exposed.
Nevertheless, enhancing mechanisms for robust privacy protection remains a central objective for our future research.

\vspace{-5pt}
\paragraph{Remark on Theoretical Effectiveness.}
\algopt builds upon well-established theoretical foundations.
In particular, RELA \cite{sun2024efficiency} has theoretically demonstrated that leveraging task-agnostic models, such as pre-trained models, can accelerate model learning.
Nevertheless, \algopt departs significantly from RELA in both its research objectives and methodological design.
Specifically, \algopt introduces a collaborative optimization approach tailored for unlabeled data, targeting a fundamentally distinct problem domain.
The challenges associated with collaborative data optimization are unique to \algopt and remain unaddressed by RELA.
Furthermore, the metric we employ to evaluate the quality of prior models, uniform value loss \citep{wang2020understanding}, has been theoretically validated for its effectiveness.


\vspace{-5pt}
\section{Experiments}
\vspace{-5pt}
We conduct extensive experiments to demonstrate the key advantages of \algopt in \secref{sec:exp_sota}. 
Specifically, \textit{first}, to evaluate its efficacy and efficiency in utilizing unlabeled data, we compare \algopt with state-of-the-art self-supervised learning methods.
\textit{Second}, to examine the necessity of distributed optimization, we further compare \algopt with centralized optimization approaches.
\textit{Third,} we train diverse model architectures on the optimized data, thereby assessing its generalizability and reusability. 
\textit{Forth,} we demonstrate the potential of \algopt for continuous data optimization, showing how it continuously enhances data quality. 
\textit{Furthermore,} we explore factors that influence its effectiveness (\secref{sec:exp_factor}), including different prior datasets and prior models. 
\textit{Finally,} we present comprehensive ablation studies (\secref{sec:exp_ablation}) to verify the impact of each module in \algopt and its lightweight design.

\vspace{-5pt}
\subsection{Experimental Setup}\label{sec:exp_setup}

\paragraph{Datasets and Networks:}
We conduct experiments on both large-scale and small-scale datasets, including ImageNet-1k (224 $\times$ 224) \citep{deng2009imagenet},
Tiny-ImageNet (64 $\times$ 64) \citep{le2015tiny}, CIFAR-100 \citep{krizhevsky2009learning} and CIFAR-10 (32 $\times$ 32) \citep{krizhevsky2009cifar}.
Following previous self-supervised studies \citep{he2020momentum,chen2020simple, grill2020bootstrap,chen2021exploring,assran2023self,zhang2024self}, we employ a range of backbone architectures to evaluate the generalizability of our method, including ResNet-\{18, 50, 101\} \citep{he2016deep}, ViT \citep{dosovitskiy2020image}, and a series of CLIP-based models \citep{radford2021learning}.

\vspace{-5pt}
\paragraph{Baselines:}
For the unlabeled data, following a widely used benchmark \citep{da2022solo}, we compare against state-of-the-art self-supervised (SSL) methods, including: SimCLR \citep{chen2020simple}, BYOL \citep{grill2020bootstrap}, DINO \citep{caron2021emerging}, MoCo \citep{he2020momentum}, SimSiam \citep{chen2021exploring}, SwAV \citep{caron2020unsupervised}, and DCL \citep{yeh2022decoupled}.
Notably, we do not compare with knowledge distillation (KD) or dataset distillation (DD) methods, since the training paradigm of KD differs significantly from ours, while DD primarily focuses on the labeled data.


\vspace{-5pt}
\paragraph{Evaluation and Metrics:}
Following previous benchmarks \citep{grill2020bootstrap,chen2021exploring}, we evaluate the representation quality of models by evaluating their test accuracy (\%) using an offline linear probing strategy.
Additionally, computational efficiency quantified by time cost (s).

\vspace{-5pt}
\paragraph{Implementation Details:}
The proposed algorithm, \algopt, involves an open data platform and facilitates interaction among multiple participants.
The implementation follows five key steps (detailed in \secref{sec:framework}), and more details are provided in \appref{app:details}.
\begin{enumerate}   [nosep, leftmargin=16pt, itemsep=3pt]
    \item For step \circled{1}: the training dataset is evenly distributed among all participants.
    \item For steps \circled{2} and \circled{3}, in practical applications, \textit{each participant can use publicly pre-trained models or their own models directly as the prior model.}
          To simulate the diversity of prior models in practical applications, we use a series of pre-trained CLIP-based models.
    \item Steps \circled{4} and \circled{5} involve uploading data to the open platform for aggregation.
          Subsequently, for training on optimized data, we use the AdamW optimizer, the same as baselines.
          The size of mini-batch is set as $128$, except for ImageNet-1K, where a mini-batch size of 256 is utilized. 
\end{enumerate}   
All experiments are conducted using 4 NVIDIA RTX 4090 GPUs.
For all experiments, we utilize 3 random seeds and report both the mean and variance of the results.
For fair comparisons, all methods in the experiments are executed with the same hyperparameters.



%


\vspace{-5pt}
\subsection{What are the advantages of \algopt?}\label{sec:exp_sota}

\begin{table*}[!t]
    \centering
    \caption{\textbf{Comparison of \algopt with Self-Supervised Learning Methods in Accuracy (\%) and Training Time (s).}
        We use four datasets: CF-10 (CIFAR-10), CF-100 (CIFAR-100), T-IN (Tiny-ImageNet), and IN-1K (ImageNet-1K).
        The best results are marked in \textbf{bold}.
        $\uparrow$ means the \textit{performance} improvement over the \underline{second-best} result.
        $\times$ denotes the factor of \textit{training speed} compared to the second-best result.}
        \vspace{-5pt}
    \label{tab:sota_all}
    \resizebox{\textwidth}{!}{
        \begin{tabular}{@{}l|c|cccccccc@{}}
            \toprule
            Dataset                 & Metric    & BYOL           & DINO            & MoCo           & SimCLR         & SimSiam        & SwAV           &
            DCL                     & \algopt (Ours)                                                                                                                                                                                                         \\ \midrule
            \multirow{3}{*}{CF-10}  & Acc. (\%) & 82.8 $\pm$ 0.1 & 82.6 $\pm$ 0.0  & 82.9 $\pm$ 0.1 & 83.1 $\pm$ 0.0 & 79.0 $\pm$ 0.0 & 82.9 $\pm$ 0.1 & \underline{83.9 $\pm$ 0.1} & \cellcolor[HTML]{f0f0f0}\textbf{89.5 $\pm$ 0.1 ($\uparrow$ 5.6)}            \\
                                    & Time (s)  & 1{,}376.56     & 1{,}457.22      & 1{,}349.56     & 1{,}114.81     & 1{,}090.79     & 1{,}012.74     & 1{,}783.34     & \cellcolor[HTML]{f0f0f0}\textbf{540.43 ($\times$ 1.87)}                     \\
            \midrule
            \multirow{3}{*}{CF-100} & Acc. (\%) & 51.7 $\pm$ 0.1 & 51.0 $\pm$ 0.0  & 57.8 $\pm$ 0.1 & 55.4 $\pm$ 0.0 & 44.6 $\pm$ 0.1 & 53.2 $\pm$ 0.1 & \underline{58.2 $\pm$ 0.2} & \cellcolor[HTML]{f0f0f0}\textbf{67.3 $\pm$ 0.1 ($\uparrow$ 9.1)}           \\
                                    & Time (s)  & 1{,}406.17     & 1{,}419.69      & 1{,}425.80     & 1{,}103.45     & 1{,}139.14     & 1{,}072.44     & 1{,}701.49     & \cellcolor[HTML]{f0f0f0}\textbf{548.11 ($\times$ 1.95)}                     \\
            \midrule

            \multirow{3}{*}{T-IN}   & Acc. (\%) & 43.9 $\pm$ 0.2 & 36.1 $\pm$ 0.0  & 42.4 $\pm$ 0.2 & 41.5 $\pm$ 0.1 & 40.8 $\pm$ 0.0 & 39.9 $\pm$ 0.1 & \underline{44.6 $\pm$ 0.0} & \cellcolor[HTML]{f0f0f0}\textbf{60.3 $\pm$ 0.1} \textbf{($\uparrow$ 15.7)} \\
                                    & Time (s)  & 7{,}086.62     & 7{,}030.90      & 7{,}133.98     & 5{,}621.33     & 5{,}531.92     & 5{,}540.96     & 9{,}201.51     & \cellcolor[HTML]{f0f0f0}\textbf{2{,}852.67 ($\times$ 1.94)}                 \\
            \midrule
    
            \multirow{3}{*}{IN-1K}  & Acc.(\%)  & \underline{61.9 $\pm$ 0.1} & 52.2  $\pm$ 0.0 & 57.6 $\pm$ 0.0 & 58.0 $\pm$ 0.0 & 55.8 $\pm$ 0.1 & 57.2 $\pm$ 0.1 & 60.6 $\pm$ 0.1 & \cellcolor[HTML]{f0f0f0}\textbf{69.8 $\pm$ 0.1} \textbf{($\uparrow$ 7.9)}  \\
                                    & Time (s)  & 133{,}766.19   & 133{,}156.88    & 150{,}420.36   & 99{,}176.29    & 98{,}656.57    & 96{,}134.98    & 102{,}450.84   & \cellcolor[HTML]{f0f0f0}\textbf{80{,}096.43 ($\times$ 1.20)}                \\
    
            \bottomrule
        \end{tabular}
        \vspace{-25pt}
    }

\end{table*}

\paragraph{Comparison with SSL Methods.}
As shown in \tabref{tab:sota_all}, \textit{our \algopt demonstrates superior performance and efficiency compared to existing self-supervised learning methods.}
We also visualize the training dynamic in \figref{fig:training_curves}.
Specifically, \algopt achieves an improvement of 7.9\% over the leading self-supervised approach BYOL on ImageNet-1K.
For Efficiency, \algopt demonstrates a substantial improvememt across various datasets.
Notably, on the Tiny-ImageNet, \algopt achieves a training speed that surpasses the efficient method SimSiam by a factor of approximately $\times 1.94$.

\begin{wraptable}{R}{0.3\textwidth}
    \vspace{-13pt}
    \centering
    \caption{Comparison with Centralized Optimization.}
    \vspace{-10pt}
    \label{tab:centralized-cifar100}
    \resizebox{.28\textwidth}{!}{
        \begin{tabular}{lcc}
            \toprule
            Method & Time (s) & Acc. (\%) \\
            \midrule
            Centralized & 23.71 & 62.1 $\pm$ 0.1 \\
            \cellcolor[HTML]{f0f0f0} Ours        & \cellcolor[HTML]{f0f0f0}\textbf{16.31} & \cellcolor[HTML]{f0f0f0}\textbf{65.8 $\pm$ 0.1} \\
            \bottomrule
            \end{tabular}}
        \vspace{-12pt}
    \end{wraptable}
    
\vspace{-5pt}
\paragraph{Comparison with Centralized Optimization.}
A key advantage of \algopt lies in its ability to use diverse prior models in parallel, thereby enabling efficient optimization.
To validate this, we compare \algopt with centralized optimization, where a single model is used to optimize all unlabeled data.
Specifically, we consider 10 prior models of varying quality on CIFAR-100.
For the centralized setting, we report the mean performance of these 10 independently selected models to ensure fairness.
As summarized in \tabref{tab:centralized-cifar100}, \textit{\algopt consistently demonstrates superior efficiency and efficacy}, verifying the benefits of distributed over centralized optimization.
This improvement arises from the proposed target alignment strategy (\secref{sec:uniform_value}), which leverages high-quality priors to enhance the target distribution of weaker models.

While one might envision an ideal centralized solution that exclusively employs the best prior model, such an approach is rarely practical in real-world scenarios
A major concern is fairness and computational burden, as concentrating all computation on a single party imposes excessive cost and discourages participation.
Another challenge is privacy, since high-performing models are typically proprietary, and centralizing them may violate ownership or data-sharing constraints.
In contrast, \algopt collaboratively leverage diverse prior models, achieving competitive performance while substantially reducing the burden on any individual participant.

\begin{wraptable}{R}{0.35\textwidth}
    \vspace{-13pt}
    \centering
    \caption{Comparison of \algopt with BYOL on Diverse Networks.}
    \label{tab:app_sota_network}
    \vspace{-10pt}
    \resizebox{.35\textwidth}{!}{
        \begin{tabular}{c|cc}
            \toprule
                   & \multicolumn{2}{c}{Method}                        \\ \midrule
            Network & BYOL                             & \algopt               \\ \midrule
            ResNet-50 &  60.4 $\pm$ 0.1                              & \textbf{63.8 $\pm$ 0.0}  \\
            ResNet-101 &  61.5 $\pm$ 0.2                              & \textbf{65.7 $\pm$ 0.2}          \\
            MobileNet-v2 & 24.0 $\pm$ 0.5                    & \textbf{58.1 $\pm$ 0.0} \\
            Efficientnet-b0 & 2.3 $\pm$ 0.2                 & \textbf{70.7 $\pm$ 0.2} \\
            ViT &  38.5 $\pm$ 0.1                           & \textbf{57.8 $\pm$ 0.1} \\
            \bottomrule
        \end{tabular}}
        \vspace{-12pt}
\end{wraptable}

\vspace{-5pt}
\paragraph{Generalizability and Reusability of Optimized Data.}\label{sec:exp_reusability}
Another key advantage of our optimized data lies in its strong generalizability and reusability: once constructed, it can be directly employed for downstream diverse model training without further modification.
To evaluate this advantage, we conduct experiments by training a variety of neural architectures on the optimized data and compare the results to a strong baseline, BYOL, which relies on training from scratch on the original unlabeled dataset.
The results are summarized in \tabref{tab:app_sota_network}.

Obviously, \algopt consistently delivers \textit{significant performance improvements over BYOL across multiple architectures.}
In particular, BYOL suffers substantial degradation when applied to lightweight networks such as MobileNet-v2.
A plausible explanation is that these models are more sensitive to unstable batch normalization (BN) statistics in early network layers.
Instead, our optimized data exhibits \textit{strong generalization} across diverse architectures.

\begin{wrapfigure}{r}{0.3\textwidth}
    \vspace{-1.5em}
    \centering
    \includegraphics[width=.3\textwidth]{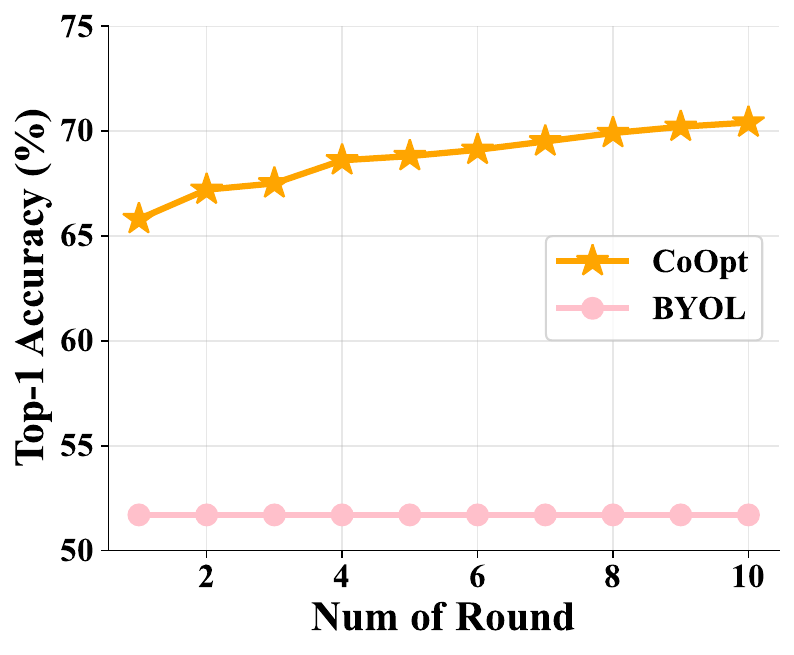}
    \vspace{-15pt}
    \caption{
        A Practical Scenario: Continuous Optimization.
    }
    \label{fig:continuous_opt}
    \vspace{-1.3em}
\end{wrapfigure}

\vspace{-5pt}
\paragraph{Continuous Data Optimization.}\label{sec:exp_continuous}
We further explore a practical scenario where the prior models undergo temporal evolution.
For example, a participant's initial model, such as ResNet-50, might be upgraded to a higher-capacity model like ResNet-101 as their computational resources improve. Consequently, the process can be treated as a dynamic and continuous procedure.
The detailed description of the process is provided in \appref{app:continuous}.

We simulate this scenario by making random $20$\% of the participants increase their model capacity in each round.
The training curves across 10 rounds on CIFAR-100 are shown in \figref{fig:continuous_opt}.
The results demonstrate that \textit{in \algopt, as the prior models evolve, the quality of the targets improves, thereby facilitating continuous optimization.}
In particular, over 10 rounds, the continuous optimization setting yields a 4.6\% performance gain.

\subsection{What influence the effectiveness of \algopt?}\label{sec:exp_factor}
We investigate the key factors that influence the effectiveness of the optimized data. 
Since the optimization of unlabeled data in \algopt relies solely on task-agnostic prior models, 
we focus on two primary aspects in these experiments: prior datasets and prior models.

\begin{figure*}[!t]
    \begin{subfigure}{0.24\textwidth}
        \centering
        \includegraphics[width=\textwidth]{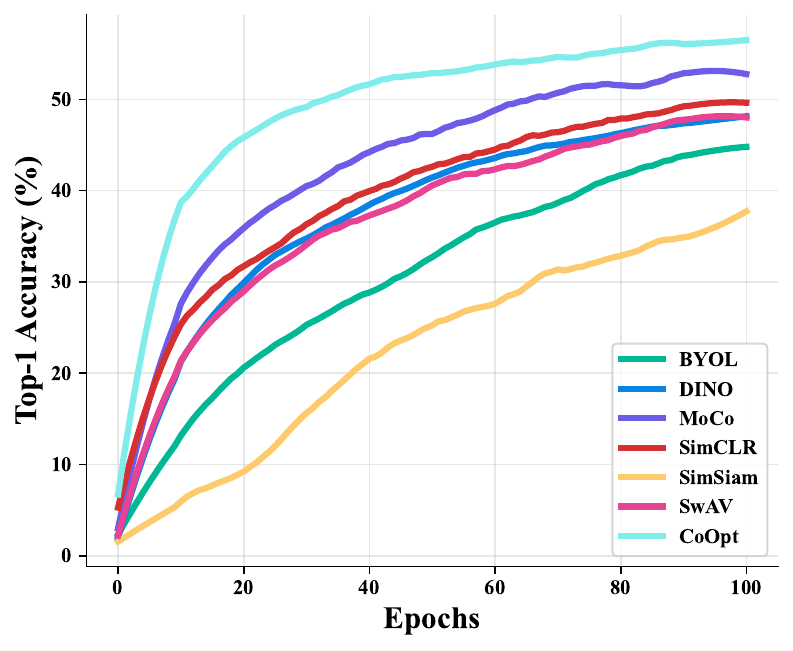}
        \caption{Training Curves}
        \vspace{-5pt}
        \label{fig:training_curves}
    \end{subfigure}
    \begin{subfigure}{0.24\textwidth}
        \centering
        \includegraphics[width=\textwidth]{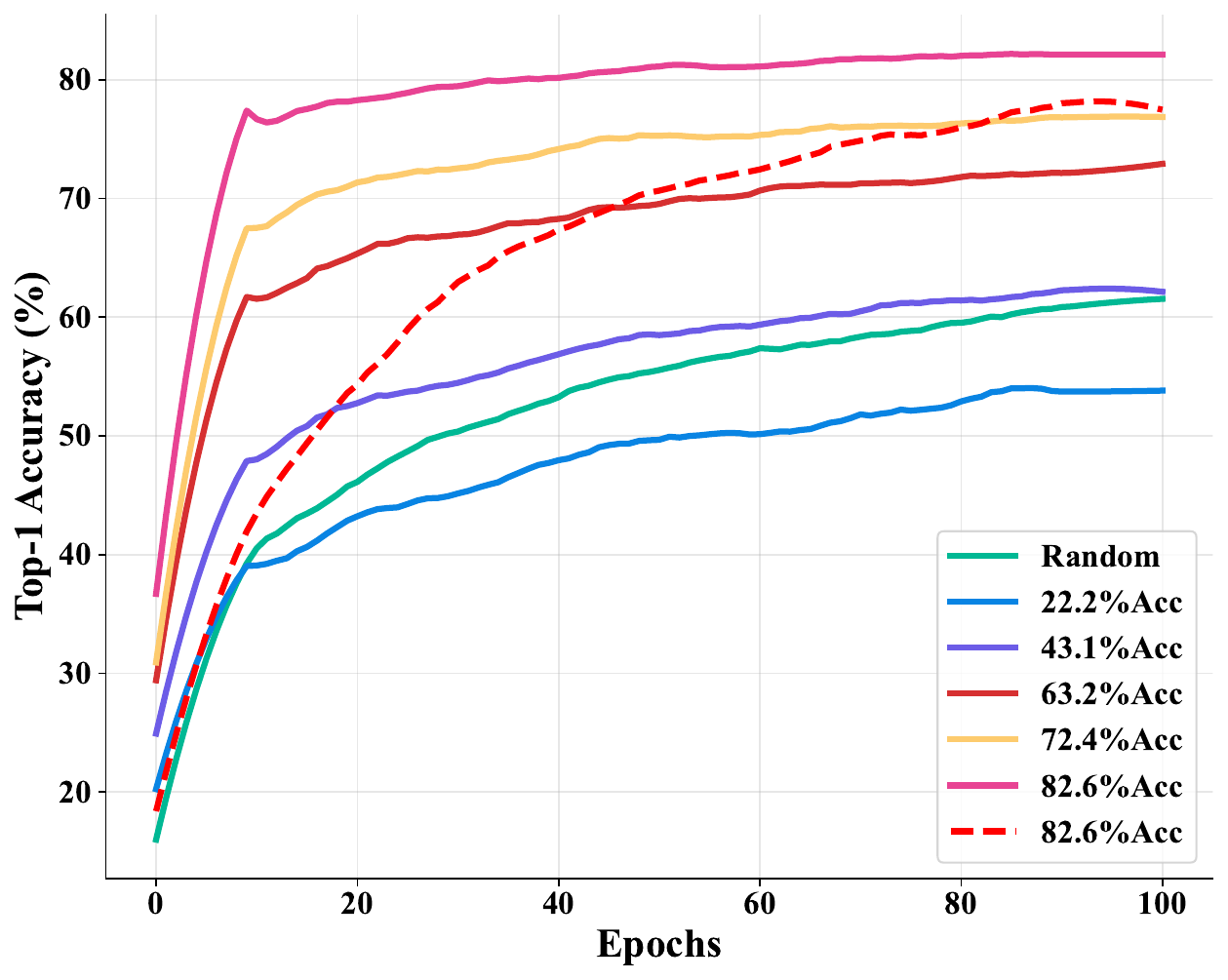}
        \caption{
            Varying Prior Models}
        \vspace{-5pt}
        \label{fig:weak_model}
    \end{subfigure}
    \begin{subfigure}{0.24\textwidth}
        \centering
        \includegraphics[width=\textwidth]{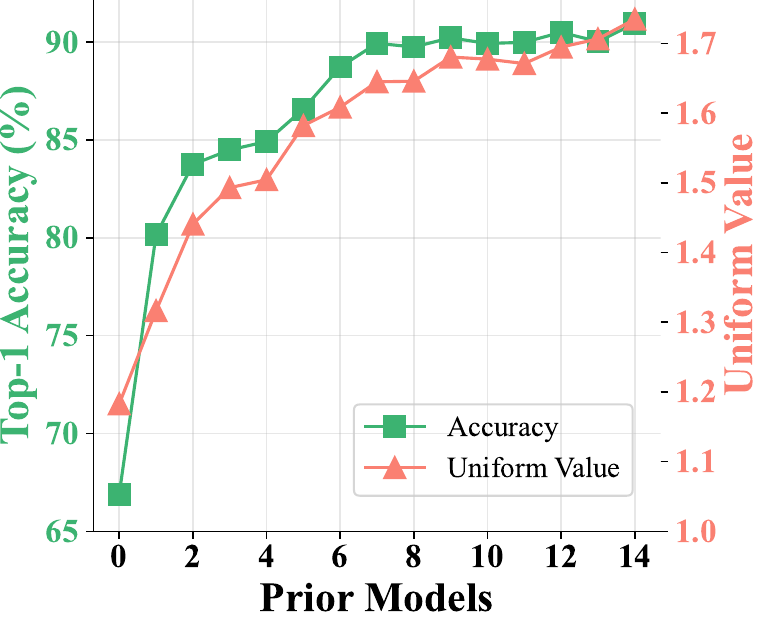}
        \caption{
            Uniform Value
        }
        \vspace{-5pt}
        \label{fig:ablation_uniform}
    \end{subfigure}
    \begin{subfigure}{0.24\textwidth}
        \centering
        \includegraphics[width=\textwidth]{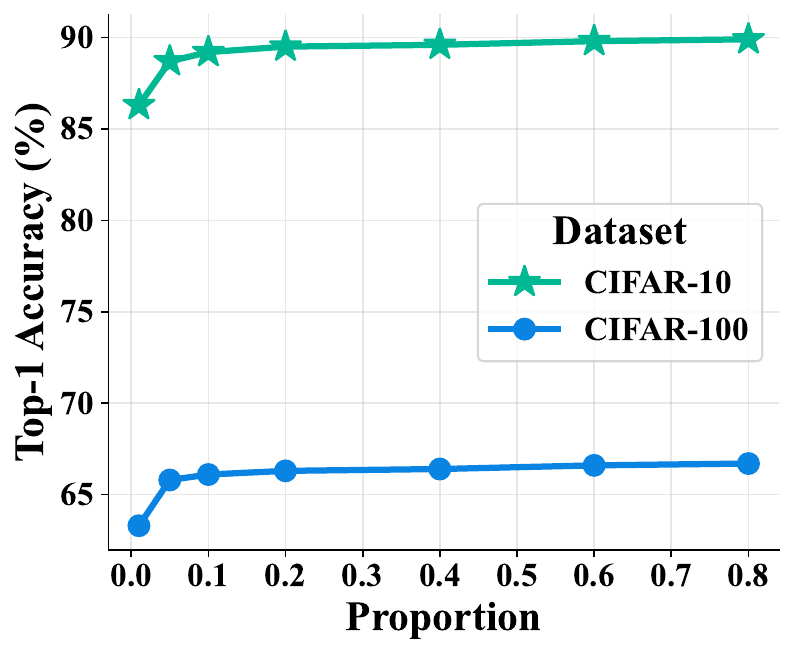}
        \caption{
            Shared Data Size
        }
        \vspace{-5pt}
        \label{fig:ablation_size}
    \end{subfigure}
    \caption{\textbf{Comprehensive Analysis of \algopt.}
        \textit{\textbf{(a)} Training curves:} Comparison of SSL methods and our \algopt.
        \textit{\textbf{(b)} Prior Models With Varying Accuracies.}
        Even with a very weak prior model, \algopt accelerates the early-stage training.
        \textit{\textbf{(c)} Correlation Verification:} Verify the correlation between the uniform value and performance.
        \textit{\textbf{(d)} Influence of shared data size:}
        As shared data's size increases, the performance gains diminish.
    }
    \label{fig:xxx}
\end{figure*}

\begin{table*}[!t]
    \centering
    \caption{\textbf{Comparison of \algopt with BYOL Across Diverse Prior Datasets.}
        For instance, ``CIFAR-10 (P)'' indicates participants' prior models are trained on CIFAR-10.
        \textbf{Bold} means the best results.
        \underline{Underline} indicates the results when the prior dataset is identical to the training data.
        All models are based on ResNet-18.
    }
    \vspace{-5pt}
    \label{tab:sota_prior_dataset}
    \resizebox{\textwidth}{!}{
        \begin{tabular}{@{}llllll@{}}
            \toprule
            \multirow{2}{*}{Dataset}       & \multirow{2}{*}{\begin{tabular}[c]{@{}c@{}}BYOL\\ (Baseline)\end{tabular}} & \multicolumn{4}{c}{Our \algopt (Diverse Prior Datasets)}                                                                                                                                                    \\ \cmidrule(lr){3-6}
                                           &                                                                            & \multicolumn{1}{c}{CIFAR-10 (P)}                         & \multicolumn{1}{c}{CIFAR-100 (P)}           & \multicolumn{1}{c}{Tiny-ImageNet (P)}       & \multicolumn{1}{c}{ImageNet-1K (P)}                  \\ \midrule
            \multirow{1}{*}{CIFAR-10}      & 82.8 $\pm$ 0.1                                                             & \underline{86.6 $\pm$ 0.0 ($\uparrow 3.8$)}              & 80.9 $\pm$ 0.0 ($\downarrow 1.9$)           & 81.6 $\pm$ 0.1 ($\downarrow 1.2$)           & \textbf{88.1 $\pm$ 0.0 ($\uparrow$ 5.3)}             \\
            \multirow{1}{*}{CIFAR-100}     & 51.7 $\pm$ 0.1                                                             & 54.9 $\pm$ 0.1 ($\uparrow 3.2$)                          & \underline{60.0 $\pm$ 0.1 ($\uparrow 8.3$)} & 56.8 $\pm$ 0.0 ($\uparrow 5.1$)             & \textbf{63.7 $\pm$ 0.0 ($\uparrow$ 12.0)}            \\
            \multirow{1}{*}{Tiny-ImageNet} & 43.9 $\pm$ 0.2                                                             & 38.3 $\pm$ 0.0 ($\downarrow 5.6$)                        & 40.2 $\pm$ 0.1 ($\downarrow 3.7$)           & \underline{49.0 $\pm$ 0.0 ($\uparrow 5.1$)} & \textbf{55.8 $\pm$ 0.1 ($\uparrow$ 11.9)}            \\
            \multirow{1}{*}{ImageNet-1K}   & 61.9 $\pm$ 0.1                                                             & 31.7 $\pm$ 0.1 ($\downarrow 30.2$)                       & 31.8 $\pm$ 0.0 ($\downarrow 30.1$)          & 40.5 $\pm$ 0.0 ($\downarrow 21.4$)          & \textbf{\underline{71.2 $\pm$ 0.0 ($\uparrow$ 9.3)}} \\\bottomrule
        \end{tabular}}
    \vspace{-10pt}
\end{table*}

\paragraph{Influence of Prior Datasets.}
To rigorously evaluate the influence of prior datasets that are used for training prior models, we perform an analysis across scenarios where the prior datasets used for prior models either align with or differ from the unlabeled training dataset.
For instance, in the aligned scenario, the training dataset is CIFAR-10, and the prior models are also trained on CIFAR-10 (CIFAR-10 (P)).
Conversely, in the divergent scenario, the training dataset remains CIFAR-10, while the prior models are trained on CIFAR-100 (P).
Details of these models are provided in \appref{app:prior_dataset}.
We evaluate our approach on four publicly available datasets, with the results summarized in \tabref{tab:sota_prior_dataset}.

Obviously, for \textit{all} unlabeled training datasets, \textit{employing prior models trained on ImageNet-1K consistently yields notable performance gains,} attributed to their strong generalization abilities.
This observation is particularly relevant in practical applications, as most publicly available pre-trained models are derived from ImageNet-1K or even larger datasets. 
On the other hand, for complex training datasets, leveraging prior models trained on simpler datasets may result in degraded performance compared to BYOL. 
This is likely due to the limited informativeness of simpler prior datasets, which provide weaker guidance. 
We further examine the influence of weak models in \figref{fig:weak_model}.

\begin{wraptable}{R}{0.35\textwidth}
    \vspace{-12pt}
    \centering
    \caption{Comparison of \algopt with BYOL in Presence of Human or Weak Prior Models.}
    \vspace{-10pt}
    \label{tab:sota_human_random}
    \resizebox{.35\textwidth}{!}{
        \begin{tabular}{c|cc|c}
            \toprule
                   & \multicolumn{2}{c|}{Prior Models} & Dataset                             \\ \midrule
            Method & Human                             & Weak      & CIFAR-10                \\ \midrule
            BYOL   & --                                & --        & 82.8 $\pm$ 0.1          \\ \midrule
            \multirow{3}{*}{\algopt}
                   & \ding{55}                         & \ding{55} & 89.5 $\pm$ 0.1          \\
                   & \ding{55}                         & \ding{51} & 89.2 $\pm$ 0.2          \\
                   & \ding{51}                         & \ding{55} & \textbf{90.5 $\pm$ 0.1} \\
                   & \ding{51}                         & \ding{51} & \underline{89.8 $\pm$ 0.1}          \\
            \bottomrule
        \end{tabular}}
        \vspace{-5pt}
\end{wraptable}

\vspace{-3pt}
\paragraph{Special Cases of Prior Models: Human or Weak Involvement.}
In real-world applications, extreme cases arise due to the varying capabilities of participants.
For example, some participants have extensive resources and can employ human annotators for labeling, while others may have limited resources and rely on weak models with inferior generalization abilities.
In this experiment, we define weak models as those trained during intermediate stages that are even far from convergence.
To simulate the conditions, in addition to the prior models used in the first experiment, we incorporate $5$ prior models, either human or weak models to the data optimization process.
The results are summarized in \tabref{tab:sota_human_random}.
Surprisingly, even the inclusion of weaker models contributes to enhancing the final performance, indicating that such models can still provide valuable information.
Moreover, it is important to note that \textit{the integration of high-capacity, human-like models results in significant performance improvements.}

\begin{figure}[!t]
    \centering
    \begin{subfigure}{0.235\textwidth}
        \includegraphics[width=\linewidth]{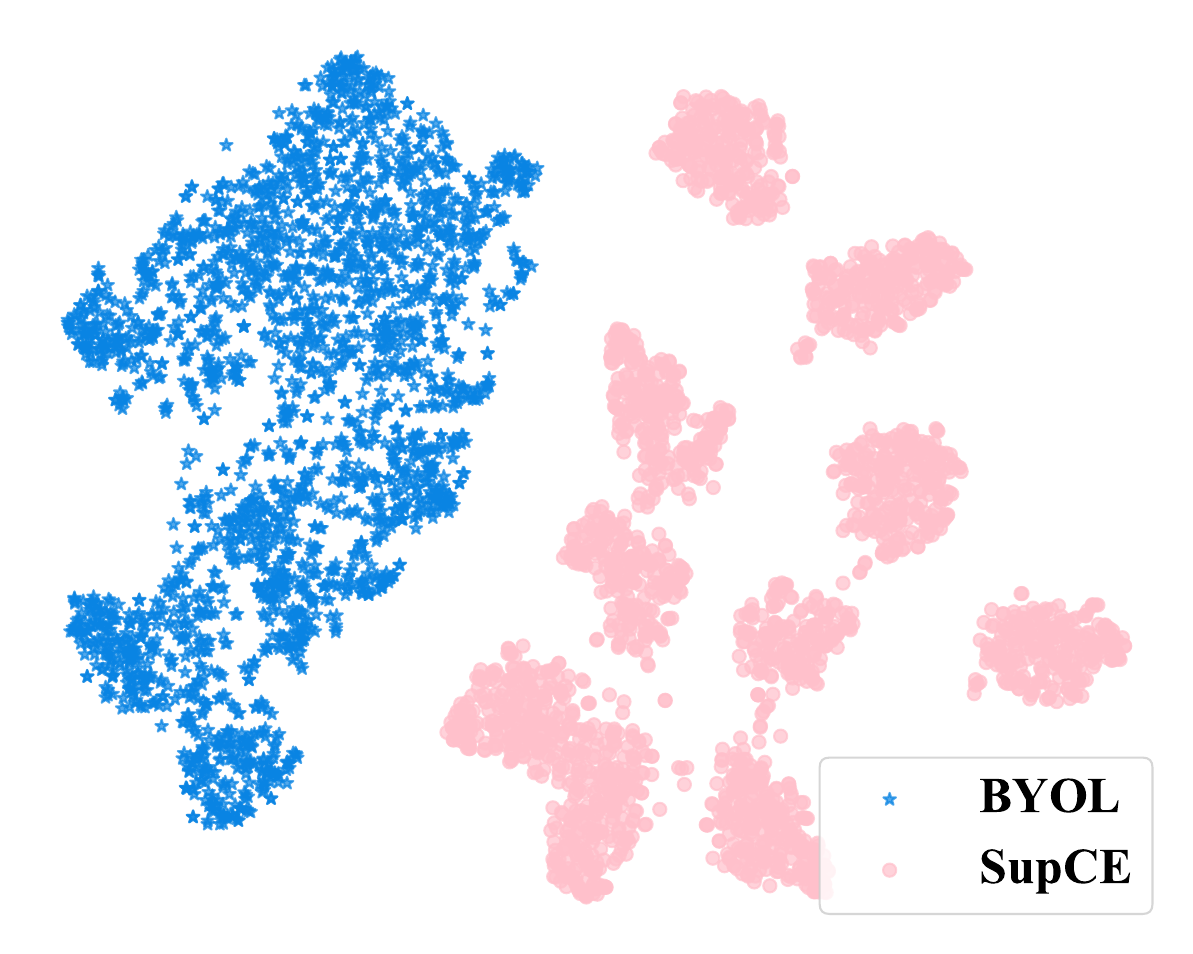}
        \caption{Before Alignment}
        \label{fig:vis_inconsistency_before}
        \vspace{-5pt}
    \end{subfigure}
    \begin{subfigure}{0.235\textwidth}
        \includegraphics[width=\linewidth]{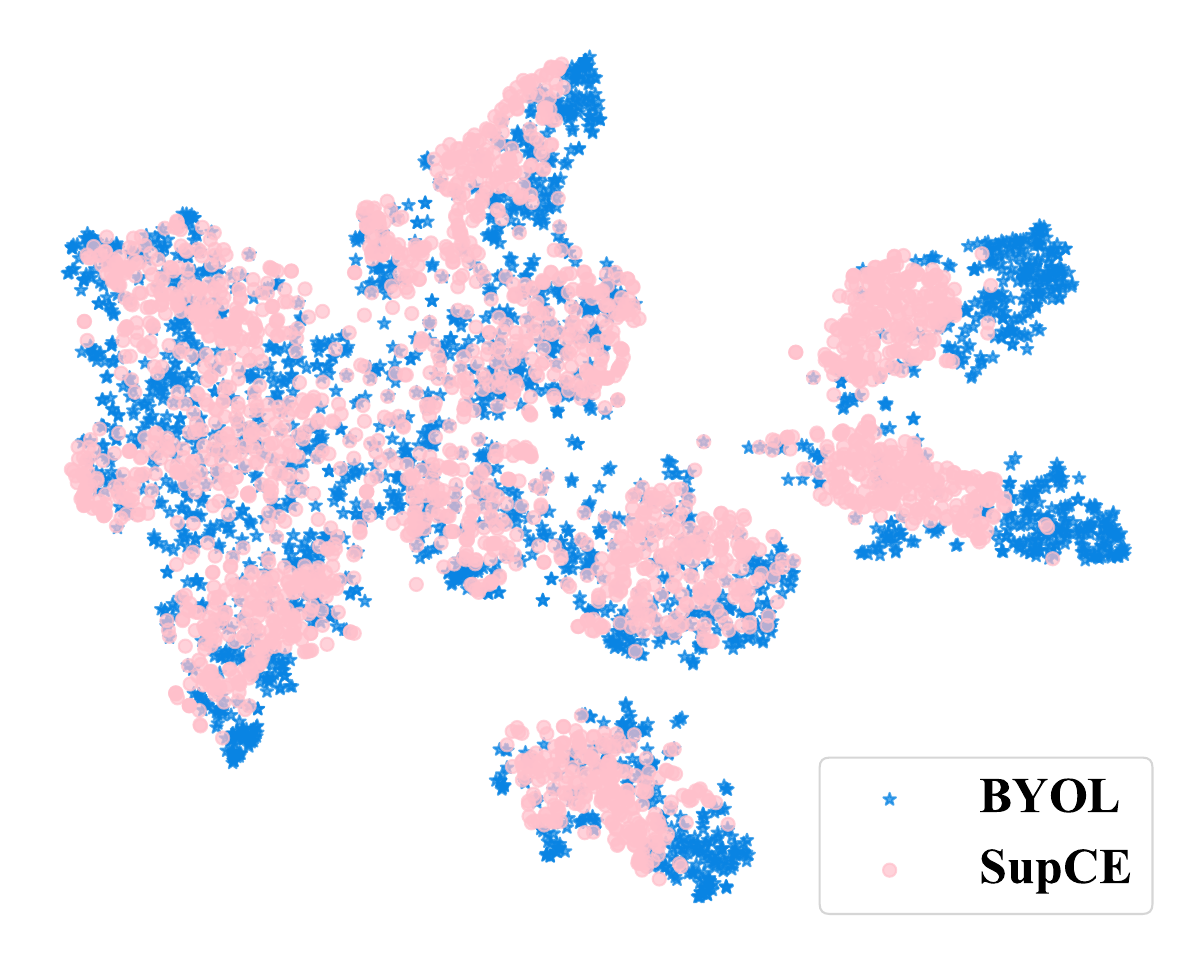}
        \caption{Align (BYOL $\rightarrow$ CE)}
        \label{fig:vis_inconsistency_after_bc}
        \vspace{-5pt}
    \end{subfigure}
    \begin{subfigure}{0.235\textwidth}
        \centering
        \includegraphics[width=\textwidth]{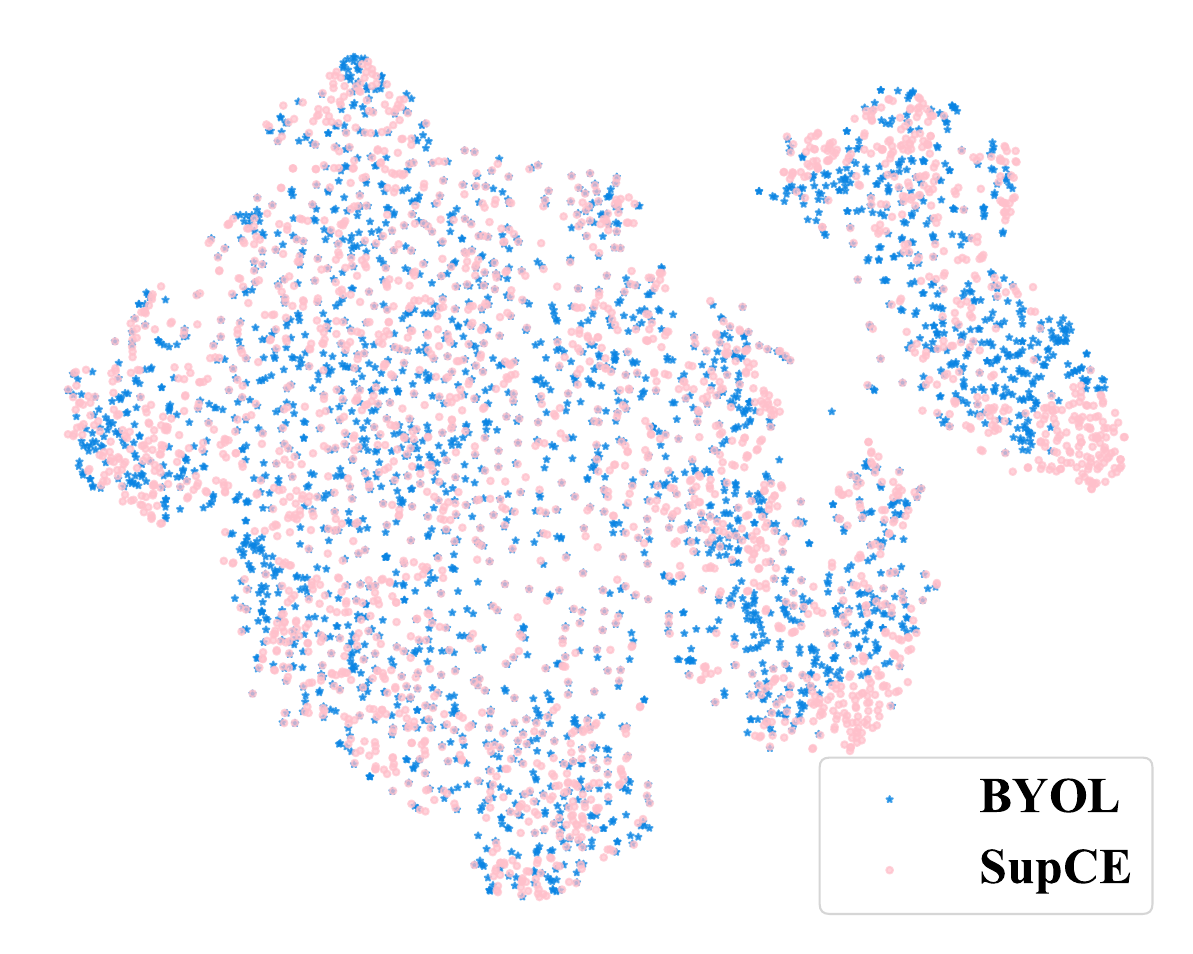}
        \caption{Align (CE $\rightarrow$ BYOL)}
        \label{fig:vis_inconsistency_after_cb}
        \vspace{-5pt}
    \end{subfigure}
    \begin{subfigure}{0.24\textwidth}
        \includegraphics[width=\linewidth]{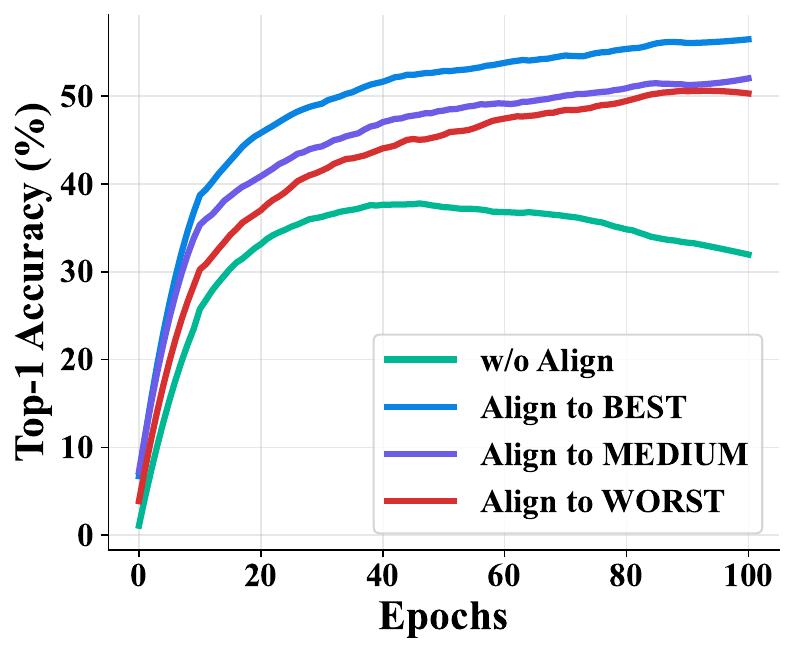}
        \caption{Training curves.}
        \label{fig:ablation_align}
        \vspace{-5pt}
    \end{subfigure}
    \caption{\textbf{Effectiveness of Target Distribution Alignment.}
        (a), (b), (c): Visualization of t-SNE for optimized targets generated by two distinct models (BYOL (acc. = 82\%) and SupCE (acc. = 90\%).)
        Aligning to the worse model (c) results in diminished target quality.
        (d): Training curves with and without alignment.
    }
    \vspace{-10pt}
    \label{fig:xxxx}
\end{figure}

\vspace{-3pt}
\paragraph{Extreme Cases of Prior Models: All are Weak.}
Moreover, we conduct experiments with only weak model, as shown in \figref{fig:weak_model}.
Here, the dashed line represents the training curve of BYOL (baseline), and the solid lines means prior models with different accuracies.
"Prior model (BYOL)" indicates the use of BYOL as the prior model, and the accuracies of the other prior models are all lower than that of BYOL.
While a stronger prior model does yield better performance, our results demonstrate that \textit{even with a moderately weak prior model with approximately 75\% accuracy, our method can still outperform the baseline.}
More importantly, even when prior models are substantially weak, \textit{\algopt still significantly outperform baseline in the \textbf{early training stages.}}

\vspace{-5pt}
\subsection{Ablation Study}\label{sec:exp_ablation}

\paragraph{Can Uniform Value Effectively Assess the Quality of Prior Models?}
To evaluate the effectiveness of uniform value in estimating the quality of prior models, we employ a diverse set of prior models and compute both their uniform values and corresponding test accuracies.
To quantify the relationship between these two measures, we adopt the Spearman rank correlation coefficient\footnote{The formula is:
    $\rho = 1 - \frac{{6 \sum d_i^2}}{{n(n^2 - 1)}}$, where \( d_i \) represents the difference between the ranks of each pair of observations and $n$ denotes the number of observations.} \( \rho \)~\citep{zar2014spearman} to quantify the association between Uniform Value and accuracy.
As illustrated in \figref{fig:ablation_uniform}, the Spearman rank correlation coefficient is $\rho=-0.9714$, \textit{indicating a strong correlation between uniform value and model performance, thereby effectively assessing the quality of prior models.}

\vspace{-5pt}
\paragraph{Why Target Distribution Alignment is Necessary?}
In real-world applications, participants often adopt diverse prior models, leading to inconsistencies in the target space (\figref{fig:vis_inconsistency_before}).
To verify the necessity of alignment, we conduct an ablation study with and without alignment, and the training curves in \figref{fig:ablation_align} show that our alignment method improves performance by 16.9\%, veifying its effectiveness.
Furthermore, We compare three strategies: aligning to the best prior model (ours), a medium-quality prior model (\textcolor{purple1}{purple line}), and the worst prior model (\textcolor{red1}{red line}).
All strategies outperform the unaligned baseline, but aligning to the best prior model yields the largest performance gains, as reported in \tabref{tab:mis_selection} of \appref{app:exp_mis_selection}.
To analyze the underlying reasons, we employ t-SNE visualization. As shown in \figref{fig:vis_inconsistency_after_bc} and \figref{fig:vis_inconsistency_after_cb}, alignment with a high-quality model enhances the representational capability of the targets, whereas alignment with a poor model diminishes it.

\paragraph{How Much Shared Unlabeled Data Is Enough?}
The shared unlabeled data $S$ is used to estimate the uniform value and compute the transformation matrix for target alignment, as detailed in \secref{sec:uniform_value}.
To explore the influence of the data size, we vary the proportion of shared data on CIFAR-10 and CIFAR-100, as shown in \figref{fig:ablation_size}.
Obviously, \textit{as the size of the shared data increases, performance gains become marginal}, indicating that a very small fraction (around 0.05\%) is sufficient for accurate estimation and alignment.
We further validate this on ImageNet-1K (\tabref{tab:shared_data_size_imagenet} in \appref{app:exp_shared_size}), where only 0.001\% of the data achieves comparable results to larger proportions.
This minimal requirement imposes negligible additional overhead, ensuring scalability to very large datasets.

\begin{wraptable}{R}{0.3\textwidth}
    \vspace{-12pt}
    \centering
    \caption{Comparison of Time (s) on ImageNet-1K.}
    \vspace{-5pt}
    \label{tab:uniform_cost}
    \resizebox{.3\textwidth}{!}{
        \begin{tabular}{lcc}
            \toprule
            BYOL & Uniform value & Alignment \\
            \midrule
            133,766.19 & 139.16 & 36.97 \\
            \bottomrule
            \end{tabular}}
        \vspace{-5pt}
\end{wraptable}

\vspace{-5pt}
\paragraph{Does Alignment Incur Significant Overhead?}
We report the computational cost of uniform value estimation and alignment in \tabref{tab:uniform_cost}, verifying that \textit{both incur only negligible cost.}
This efficiency stems from the former requiring just a single forward pass to obtain targets, while the latter involves optimizing a lightweight matrix.

\vspace{-5pt}
\section{Conclusion}
We introduce \algopt, a pioneering data-centric, parallelized, and efficient framework for collaborative optimization of unlabeled data.
This data-centric approach results in architecture-agnostic optimized data that are reusable across diverse network architectures, while simultaneously reducing the number of training iterations required, thereby enhancing overall efficiency.
Furthemore, within \algopt, we identify a critical issue: Target Distribution Inconsistency, which arises from the diversity of prior models used in data optimization.
To mitigate this, we propose a lightweight target alignment strategy.
Extensive experiments demonstrate the superior effectiveness and efficiency of the \algopt framework across diverse datasets and architectures.
One limitation is that when all prior models are extremely weak, the overall performance inevitably degrades.
As future work, we aim to develop advanced strategies to more effectively exploit optimized data derived from all extremely weak priors, with our current study verifying efficiency in the early training stage.


\clearpage

\bibliography{resources/reference}
\bibliographystyle{configuration/iclr2026_conference}

\clearpage
\appendix

\section{Related Work}\label{sec:app_related_work}

\paragraph{Self-supervised Learning: A Model-Centric Perspective.}
Self-supervised learning \citep{simclr} aims to exploit the intrinsic relationships within unlabeled data.
For example, InstDisc \citep{wu2018unsupervised} uses instance discrimination as a pretext task.
Building on this, CMC \citep{tian2020contrastive} proposes to use multiple views of an image as positive samples and take another one as the negative.
MoCo \citep{he2020momentum} significantly increases the number of negative samples but uses a simplistic strategy for selecting positive samples.
SimCLR \citep{chen2020simple} highlights the importance of hard positive sample strategies by introducing data augmentation.
Notably, BYOL \citep{grill2020bootstrap} discards negative sampling and surpasses the performance of SimCLR \citep{chen2020simple}.

\textbf{Summary.} \textit{
    They are constrained to specific architectures and incur high computational costs due to the reliance on large batch sizes or memory banks.}

\paragraph{Knowledge Distillation: Optimzing Targets.}
Knowledge distillation \citep{hinton2015distilling} employs soft labels generated by teacher models to improve the performance of a student model and expedite its training \citep{Dong2023DisWOT:}.
Many following works aim to enhance the use of soft labels for more effective knowledge transfer.
For example, WSLD \citep{zhou2021rethinking} analyzes soft labels and distributes different weights for them from a perspective of bias-variance trade-off.
DKD \citep{zhao2022decoupled} decouples the logits and assigns different weights for the target and non-target classes.
Moreover, several studies \citep{yim2017gift, Dong2023DisWOT:} demonstrate that knowledge distillation can accelerate the training process.

\paragraph{Dataset Distillation: Optimizing Both Samples and Targets.}
Dataset distillation \citep{wang2018dataset} aims to learn a compact distilled dataset that can achieve comparable performance to the original dataset with less training cost.
The majority of methods focus on optimizing images, which can be categorized into five primary approaches \citep{lei2023comprehensive}: meta-learning frameworks \citep{wang2018dataset, zhou2022dataset}, gradient matching \citep{zhao2020dataset, zhao2021dataset}, distribution matching \citep{zhao2023dataset, yin2023squeeze}, trajectory matching \citep{cazenavette2022dataset, guo2023towards}, and decoupling frameworks \citep{yin2023squeeze, sun2024efficiency}.
Recently, some studies \citep{shang2025gift, qin2024label} have shifted focus towards label distillation, aiming to obtain high-quality soft labels. This approach has demonstrated notable efficiency and effectiveness.


\section{Experimental Details} \label{app:details}
\subsection{Implementation Details}
\paragraph{Hardware Setup.}
All experiments are conducted using 4 NVIDIA RTX 4090 GPUs. For fair comparisons, all methods in the experiments are executed with the same hyperparameters.

\paragraph{Unlabeled Training Dataset Split.}
In our framework, for all experiments, the unlabeled dataset is evenly distributed among all participants.

\subsection{Diverse Pre-trained Prior Models.}\label{app:pre-trained}
We utilize 10 CLIP-based pre-trained prior models varying from small-scale to large-scale architectures, which are downloaded using the torchvision package.
Unlabeled data is equally split among participants.

\subsection{Diverse Prior Datasets}\label{app:prior_dataset}
In this set of experiments, we investigate the effectiveness of our proposed \algopt when the prior datasets are either similar to or distinct from the training datasets.
For each prior dataset, we train 4 different models to serve as prior models. Specifically:
These 4 models are trained using two paradigms (supervised and unsupervised learning) and two architectures (ResNet-18 and ViT).
For each training dataset, the data is evenly distributed among participants.
Each prior model is assigned 1/4 of the unlabeled dataset, which it optimizes independently.
The results are then aggregated on the open platform.

\subsection{Continuous Data Optimization}\label{app:continuous}
\begin{figure}[h]
    \centering
    \includegraphics[width=.5\textwidth]{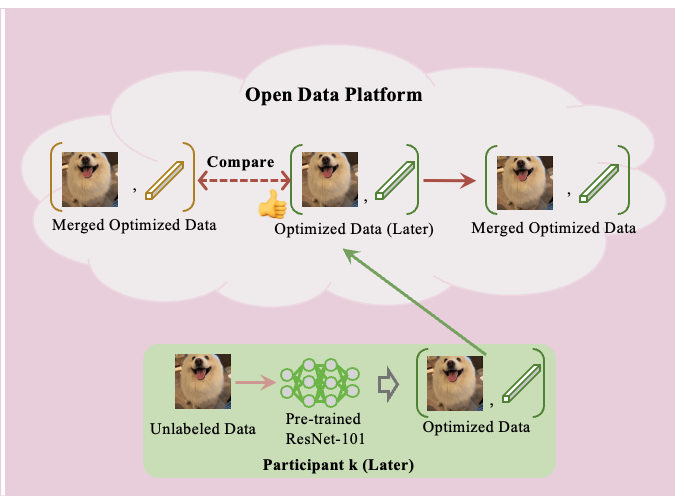}
    \caption{\textbf{Common Target Conflict Scenarios}}
    \label{fig:app_continuous}
\end{figure}

In real-world scenarios, model architectures and computational resources often evolve over time, necessitating a dynamic and continuous approach to data optimization. Unlike static optimization, where data is processed only once, this framework accommodates temporal model evolution and repeatedly refines optimized data to achieve superior results. Below, we describe the core components of this framework.
\paragraph{Dynamic Model Evolution.
}
Initially, all data is optimized using participants' current models. Over time, as computational resources improve, participants upgrade to higher-capacity architectures.
These upgraded models exhibit stronger feature extraction capabilities, enabling further refinement of previously optimized data.
This dynamic evolution transforms the optimization process into a continuous improvement cycle.
\paragraph{Open Platform for Optimized Data Comparison.}
A key feature of this framework is the inclusion of an open platform for optimized data submission and evaluation.
Participants provide newly optimized data, which is compared against previously optimized data to retain the superior results.
This comparison leverages evaluation metrics such as the uniform value of the prior model, ensuring that the dataset evolves toward higher optimization quality over successive iterations.

\paragraph{Iterative Optimization Process}
The overall process is illustrated in \figref{fig:app_continuous}.
The framework operates as a loop:
\begin{enumerate}
    \item Data is initially optimized by participants' models.
    \item As models evolve, data is re-optimized to reflect the improved capacities of the upgraded architectures.
    \item The open platform compares new and old optimization results, retaining the higher-quality data.
    \item This process repeats over multiple interaction rounds, progressively enhancing the dataset.
\end{enumerate}

\begin{table}[h]
    \centering
    \caption{Performance Under Three Selection Scenarios.}
    \label{tab:mis_selection}
    \resizebox{.8\textwidth}{!}{
        \begin{tabular}{lccccc}
    \toprule
    Dataset & BYOL & No align & Best & Medium & Worse \\
    \midrule
    CIFAR-100 & 51.7 $\pm$ 0.1 & 44.7 $\pm$ 0.0 & \textbf{65.3 $\pm$ 0.0} & 63.1 $\pm$ 0.0 & 60.1 $\pm$ 0.1 \\
    \bottomrule
    \end{tabular}}
\end{table}

\subsection{Various Alignment strageties}\label{app:exp_mis_selection}
As shown in \figref{fig:ablation_align}, even when the best prior model is mis-selected (i.e., prior models are aligned to a medium-performing prior or even the worst prior, represented by the purple and red lines, respectively), the overall performance still surpasses the “no align” baseline. Furthermore, we present the final performance in \tabref{tab:mis_selection}. The results suggest that our method continues to outperform the baseline methods under three selection scenarios and demonstrates robustness to prior model mis-selection.

\begin{table}[h]
    \centering
    \caption{Influence of Shared Data Size on Large-Scale ImageNet-1K.}
    \label{tab:shared_data_size_imagenet}
    \resizebox{.5\textwidth}{!}{\begin{tabular}{lccc}
    \toprule
    Dataset & BYOL & 0.001 & 0.1 \\
    \midrule
    ImageNet-1K & 61.9 $\pm$ 0.1 & 68.8 $\pm$ 0.1 & 68.7 $\pm$ 0.0 \\
    \bottomrule
    \end{tabular}}
    \end{table}

\subsection{Size of Shared Data}\label{app:exp_shared_size}
The shared unlabeled data $S$ is used to estimate the uniform value and compute the transformation matrix for target alignment, as detailed in \secref{sec:uniform_value}.
The shared dataset is public and randomly selected.
This eliminates privacy concerns and does not require matching any specific data distribution, thus avoiding issues of distribution mismatch or privacy leakage.
Moreover, we further evaluated our method on the large-scale ImageNet-1K. As shown in \tabref{tab:shared_data_size_imagenet}, only 0.001\% of the data samples are sufficient to achieve performance comparable to using larger amounts of data. This minimal requirement imposes negligible additional overhead, ensuring scalability to very large datasets.



\end{document}